\documentclass[conference]{IEEEtran}
\IEEEoverridecommandlockouts

\usepackage{cite}
\usepackage{amsmath,amssymb,amsfonts}
\usepackage{algorithmic}
\usepackage{graphicx}
\usepackage{textcomp}
\usepackage{xcolor}

\usepackage{tipa}
\usepackage{latexsym}
\usepackage{array}
\usepackage{booktabs}
\usepackage{colortbl}
\usepackage{multirow}
\usepackage[table]{xcolor}
\usepackage{makecell}
\usepackage{adjustbox}
\usepackage{refcount}
\usepackage{pifont}
\usepackage{eso-pic} %

\definecolor{tfRule}  {HTML}{CBD2D9}

\newcommand{\vI} {FSL1}  %
\newcommand{\vII}{FSL2}  %
\newcommand{\vIII} {FSL3}  %

\def\BibTeX{{\rm B\kern-.05em{\sc i\kern-.025em b}\kern-.08em
    T\kern-.1667em\lower.7ex\hbox{E}\kern-.125emX}}
\begin{document}

\AddToShipoutPictureBG*{%
  \AtPageLowerLeft{%
    \put(\LenToUnit{0.08\paperwidth},\LenToUnit{0.30in}){%
      \parbox[b]{0.84\paperwidth}{\scriptsize\centering
        \copyright~2026 IEEE. Personal use of this material is permitted. Permission from IEEE must be obtained for all other uses, in any current or future media, including reprinting/republishing this material for advertising or promotional purposes, creating new collective works, for resale or redistribution to servers or lists, or reuse of any copyrighted component of this work in other works.\\
        Author's version of the paper accepted at the 2026 IEEE Spoken Language Technology Workshop (SLT). It differs from the published version only by an added citation.}}}}

\title{Prominence and Boundary Strength Modulation in Speech Synthesis}

\author{
\IEEEauthorblockN{Pauline Mas\textsuperscript{*,\dag}, Kévin Vythelingum\textsuperscript{*}, Jonathan Chevelu\textsuperscript{\dag,\S}, Marion Ouédraogo\textsuperscript{*}, Olivier Rosec\textsuperscript{*}, Damien Lolive\textsuperscript{\ddag,\S}}
\IEEEauthorblockA{\textsuperscript{*}Voxygen, France \quad \textsuperscript{\dag}Univ. Rennes, IRISA, France \quad \textsuperscript{\ddag}Univ. South Brittany, IRISA, France \quad \textsuperscript{\S}CNRS, France}
\IEEEauthorblockA{name.surname@voxygen.fr \quad name.surname@irisa.fr}
}

\maketitle

\begin{abstract}
In Text-to-Speech synthesis (TTS), the prediction and control of prosody from text is a difficult challenge. 
Previous studies have shown that prosodic annotations from an oracle benefit TTS models, improve their prosodic rendering as well as their controllability, and can be predicted solely from text.
In this paper, we investigate different strategies of prosodic modulation and control in a FastSpeech2 architecture, using prominence and boundary strength as representation. 
We compare the performance of several prosody-informed TTS systems, trained with internal continuous predictors and/or data-driven labels, on a French audiobook dataset. Their controllability is tested by providing them with external prosodic guidance.
Label-based systems exhibit the best controllability in our experiments, while prosody-awareness especially improves boundary strength modulation.
\end{abstract}


\section{Introduction}
Text-to-speech (TTS) synthesis is the task of mapping text to a speech signal — a process that must resolve the ambiguities of written language while capturing the acoustic and prosodic variability that characterises natural speech.
Generating contextually appropriate prosody from text is a long-lasting challenge in TTS technologies: a given string of words admits multiple valid prosodic realizations, and selecting among them requires exploiting cues from grammatical structure, word choice, and communicative intent \cite{chien_2021_proso_nonautoreg}. 

Prosody control in TTS systems spans approaches that differ widely in the nature and interpretability of their representations, ranging from human-in-the-loop refinement~\cite{maurya_human_loop_2024}, global natural language descriptions~\cite{sigurgeirsson_using_2023, sigurgeirsson23_ssw}, to less interpretable unsupervised latent embeddings. 
Among the many possible approaches, categorical prosodic labels and continuous acoustic features stand out as conditioning solutions allowing for both automation and human-interpretability.

In their work, Diatlova and Shutov~\cite{diatlova_emospeech_2023} extend FastSpeech2 with Conditional Cross-Attention re-weighting phoneme states per emotion and an eGeMAPS predictor, outperforming global-embedding baselines in quality.  
Korotkova et al.'s contribution ~\cite{korotkova_word-level_2024} relies on Q-VAE clustering of Wav2Vec and RoBERTa features to obtain word-level prosodic labels. 
Their method yields a ten-class interpretable markup across three languages.
Both works propose prosodic representations that may prove difficult to actually manipulate, either because of their dimensionality (88-parameters eGeMAPS) or their granularity (10-class markup). 
Oriented towards emotion labeling, Luo et al.~\cite{luo_emotion-controllable_2024} condition a two-staged Tacotron2 model on soft labels and CWT-derived word prominence to achieve a controllability score (based on the Pearson Correlation Coefficient) of 0.95. Although more intuitive, the control method here is more indirect and is constrained on a set of embedded emotions.

A complementary strand focuses on direct control over interpretable acoustic features.
Lameris et al.~\cite{lameris_prosody-controllable_2023} add utterance-level F0 and speech rate control to a neural HMM, with emergent creaky voice synthesis at low F0.
Iliescu et al.~\cite{iliescu_controllable_2024} show with MICVAE that four sparse phoneme-level inputs suffice to recover a target prosodic rendition.
Chen et al.~\cite{chen_drawspeech_2025} go even further by conditioning a latent diffusion model on freehand prosodic sketches via a sketch-to-contour predictor, enabling word-level emphasis control. 

Suni et al. \cite{suni_prosodic_2020} demonstrate the positive impact of their categorical prosodic events on both the control and reproduction of prosodic patterns in TTS. 
In this experiment, the labels computed by their toolkit on audio references are introduced as an oracle in a Tacotron model. 
Developed to automatically annotate speech segments on prosodic aspects without requiring any form of supervision, their Wavelet Prosody Toolkit (WPT) \cite{SUNI_2017_wavelet} performs a continuous wavelet transform (CWT) on a combination of fundamental frequency, energy and duration features to quantify (and ultimately classify) prominence and structural boundary strength (hereafter called \textbf{prominences and boundaries}). In the resulting annotation, prominence may for instance signal emphasis on a word, while boundaries mark a delimitation in the sentence's grammatical structure \cite{mas-etal-2026-automatic}.
Closer to our setup, Stephenson et al.~\cite{stephenson22_interspeech} predict word prominence from English text with BERT to control a FastSpeech2 model, notably for contrastive focus.

Following their work, as well as the automatic prediction solution proposed by Mas et al.~\cite{mas-etal-2026-automatic} to estimate prominences and boundaries from text, we choose these events as representation for its interpretability, its pre-existing automatic annotation and estimation methods as well as the relative freedom it gives TTS in the acoustic realization of prosody. 

This work systematically compares strategies for controlling synthesized speech in French through prominences and boundaries.
More specifically, we seek to understand whether prosodic event-based representations improve the performance of a FastSpeech2 model \cite{ren2021fastspeech} with respect to natural speech, and to what extent an external guidance of the same nature, estimated solely from text, can influence the generated speech at inference time. 
Our contributions include: a) several prosody-aware FastSpeech2 models for French, b) a comparative study of the impact of continuous and discrete prosody representations in a multi-speaker scenario and c) an evaluation of prosodic controllability of these models.

Our paper is organized as follows. 
We begin by describing the TTS systems and different models we experiment on in Section~\ref{models_desc}, then detail the dataset (Section~\ref{data}) and experimental setup (Section~\ref{expe_setup}). Our results and analysis are reported in Section~\ref{eval} before our conclusions.

\section{System Description}\label{models_desc}

We develop various TTS systems based on FastSpeech2's architecture \cite{ren2021fastspeech} to incorporate prosodic events in speech synthesis through a range of configurations shown in Table~\ref{tab:ablation}. 
Only the baseline is prominence- and boundaries-agnostic (see Section~\ref{fs_base}), while the remaining adaptations rely on them as continuous representations, discrete label sequences or both. 
Regardless of the prosodic configuration, all of the systems compared in this work rely on a BigVGAN vocoder \cite{lee2023bigvganuniversalneuralvocoder} to convert the decoded frames into audio.

\definecolor{yes}       {HTML}{1A7A4A}   %
\definecolor{no}        {HTML}{6C717A}   %
\definecolor{rowbaseline}{HTML}{F3F4F6}  %
\definecolor{rowmodules}  {HTML}{FFF4E0}  %
\definecolor{rowlabels} {HTML}{E6F4EC}  %
\definecolor{rowfull}    {HTML}{EAF0FB}  %
\definecolor{darkgreen}{rgb}{0.0, 0.5, 0.0}
\definecolor{BurntOrange}{RGB}{204, 85, 0}

\newcommand{\YES}{\textcolor{darkgreen}{\ding{52}}}   %
\newcommand{\NO}{\textcolor{red}{\ding{55}}}           %
\newcommand{\MINUS}{\textcolor{BurntOrange}{\rule{0.8em}{1.5pt}}}

\begin{table}[!t]
  \centering
  \renewcommand{\arraystretch}{1.45}
  \caption{%
    Model configurations compared in our experiments.
  }
  \resizebox{\linewidth}{!}{%
  \begin{tabular}{%
        l                      %
        >{\bfseries}l          %
        cc                     %
        ccc                    %
      }

  \toprule
  \multicolumn{2}{l}{}
    & \multicolumn{2}{c}{\textbf{Predictors}}
    & \multicolumn{3}{c}{\textbf{Input Labels}} \\
  \cmidrule(lr){3-4}\cmidrule(lr){5-7}
  \textbf{ID\textsuperscript{*}} & \textbf{Model}
    & \textbf{Prom.} & \textbf{Bound.}
    & \textbf{Prom.} & \textbf{Bound.} & \textbf{Melody} \\
  \midrule
  {} & FastSpeech Base   & \NO   & \NO   & \NO   & \NO   & \YES  \\
  \rowcolor{rowbaseline}
  {} & FastSpeech prom   & \YES   & \YES  & \NO   & \NO   & \YES  \\
  \rowcolor{rowmodules}
  \vI & + Boundary labels        & \YES   & \YES   & \NO  & \YES   & \YES  \\
  \rowcolor{rowmodules}
  \vII & $-$ Melody labels & \YES & \YES   & \NO   & \YES   & \MINUS \\
  \rowcolor{rowmodules}
  \vIII & + Prominence labels        & \YES   & \YES   & \YES   & \YES  & \MINUS  \\
  \rowcolor{rowlabels}
  {} & FastSpeech labels & \NO  & \NO  & \YES   & \YES   & \NO  \\
  
  \bottomrule
  \multicolumn{2}{l}{ \YES : component included.} & %
  \multicolumn{2}{l}{ \NO : component absent.} & %
  \multicolumn{2}{l}{ \MINUS~: component removed.} & {}\\
  \multicolumn{7}{l}{\textsuperscript{*}\textit{
    \vI~= FS bound. labels, %
    \vII~= FS bound. labels (no mel.),  %
    \vIII~= FS dual-labels (no mel.) %
    }}
  \end{tabular}} \\
  \label{tab:ablation}
\end{table}

\begin{figure}[htbp]
\centerline{\includegraphics[width=\linewidth]{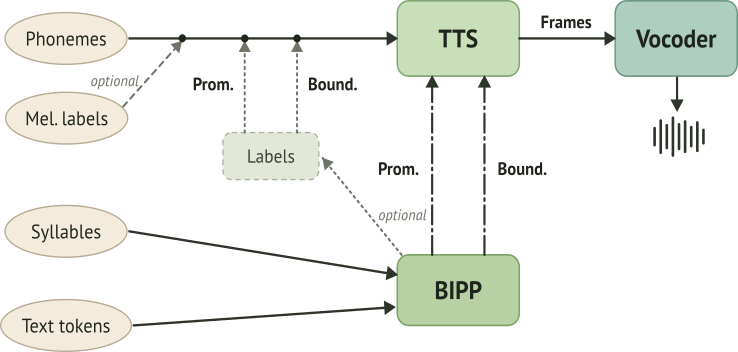}}
\caption{Overview of the experimental system. 
The prosodic events predicted from the input text by an external predictor (here BIPP \cite{mas-etal-2026-automatic}) are given to the TTS model. 
They replace the prosodic predictions from the variation predictor modules and/or the labels given alongside the input phonemes.}
\label{system_overview}
\end{figure}

\subsection{Baseline and FastSpeech Labels}\label{fs_base}

\textbf{FastSpeech2 (FS)} is a non-autoregressive encoder-decoder model that predicts audio features from phonetized sequences \cite{ren2021fastspeech}. 
Its main components include a variance adaptor that produces intermediary predictions of duration, pitch and energy at \textbf{phoneme level} forwarded to the decoder. 
While these three features need to be extracted from the ground truth signal during training, they are estimated by the variance adaptor from the encoded textual input at inference time. 

The implementation that serves us as baseline (thereafter called \textbf{FastSpeech base} or \textbf{FS base}) slightly differs from \cite{ren2021fastspeech} in that a sparse "melodic" annotation is fed to the encoder alongside phonemes. Computed from a textual analysis by an in-house tool, they are attributed to the last word of each breath group to guide its final intonation pattern.
A second implementation, based on the same architecture, is trained on prominence and boundary labels instead of melodic labels. 
Named \textbf{FastSpeech labels} (or \textbf{FS labels}), it allows us to investigate the impact of these labels as sole prosodic representation, derived either from ground-truth audio or from an external guidance (at inference time).

\subsection{FastSpeech prom}\label{fs_prom}

\begin{figure}[!tbp]
\centerline{\includegraphics[width=\linewidth]{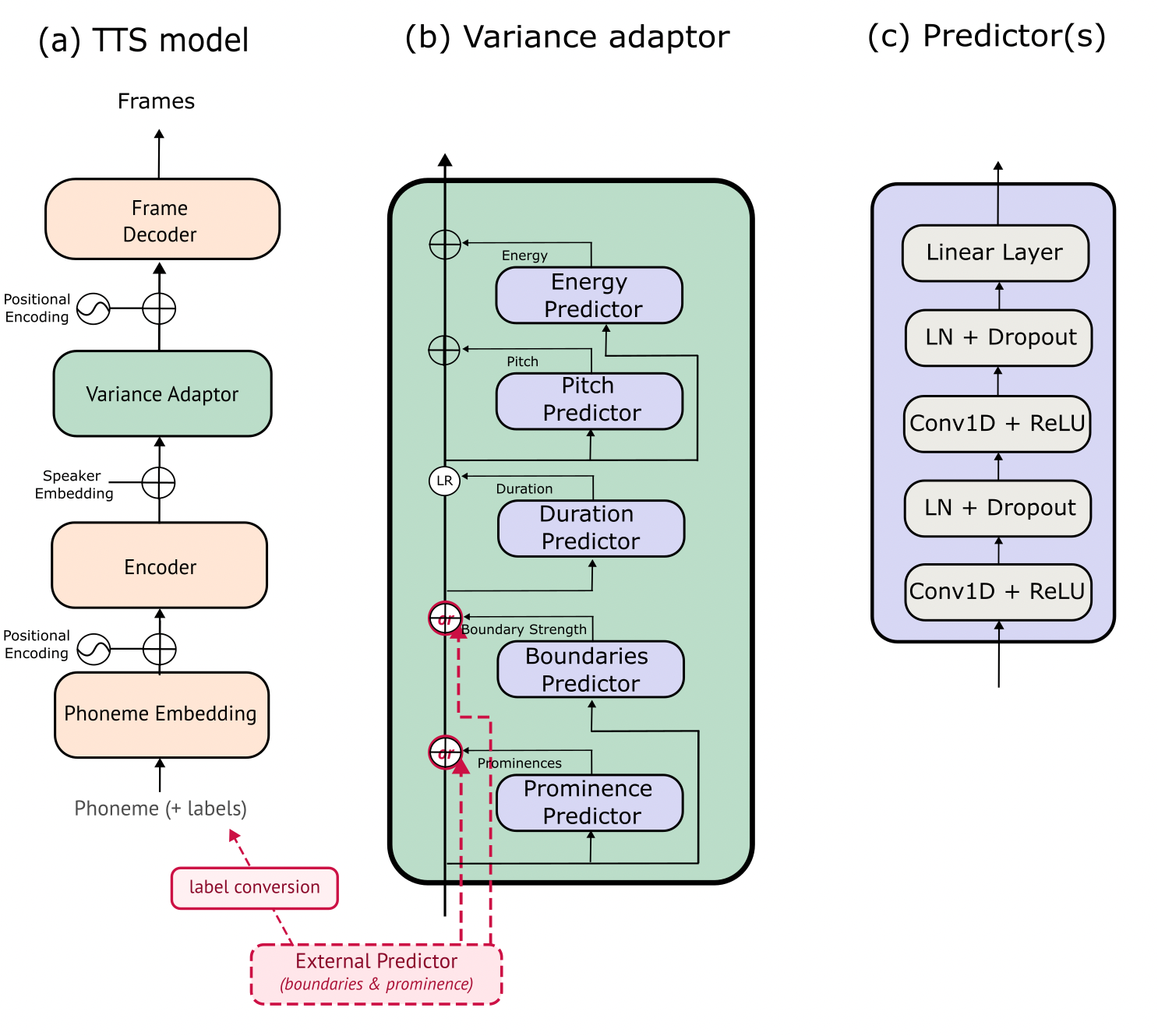}}
\caption{Architecture of the adapted FastSpeech2, with the introduced prominence and boundary predictors in the Variance Adaptor. 
At inference time, their predictions may be replaced by external predictions, as shown in red. 
}
\label{fs_labels}
\end{figure}

Following Mas et al.~\cite{mas-etal-2026-automatic}, and their assumption that predicting prominences and boundaries from text should be equivalent to a first estimation of duration, pitch and energy, we modify the architecture of our baseline to add two dedicated predictors in the variance adaptor (see Figure~\ref{fs_labels}). Positioned upstream of the duration predictor, they share its design and prediction granularity. 

By doing so, we hope prominence and boundary predictions would influence the downstream predictions within the variance adaptor and have a better impact on the overall prosody. 
This adapted architecture (hereafter called \textbf{FastSpeech prom} or \textbf{FS prom}) serves as base for the 3 variants listed in Table~\ref{tab:ablation} which differ in the input labels they receive: \textbf{FSL1} adds boundary labels; \textbf{FSL2} is \textbf{FSL1} without the melody label; \textbf{FSL3} adds both prominence and boundary labels (also without melody). All three keep the continuous predictors of \texttt{FS prom}.
At inference time, FS prom and its variants can rely either on their internal predictions or on an external prosodic event sequences, as shown in the system overview (Figure \ref{system_overview}).

\subsection{Text-based Automatic Guidance}\label{bipp_guidance}

We use and compare both inference configurations in our experiments in order to assess the extent to which the generated speech's prosody can be controlled through prosodic events guidance, as illustrated by the red lines in Figure~\ref{fs_labels}. 
Since FS is designed to rely solely on textual information at inference, we choose the \textbf{Bert-Informed Prosody Predictor (BIPP)} \cite{mas-etal-2026-automatic} as front-end guidance predictor. As the name suggests, BIPP is a transformer-based model relying on a pre-trained encoder to predict prosodic event from text.
To match with the pre-existing melodic annotation level, %
it is implemented and re-trained to predict \textbf{word-level} prominence and boundary values.

\section{Dataset}\label{data}

\begin{figure}[htbp]
\includegraphics[width=\linewidth]{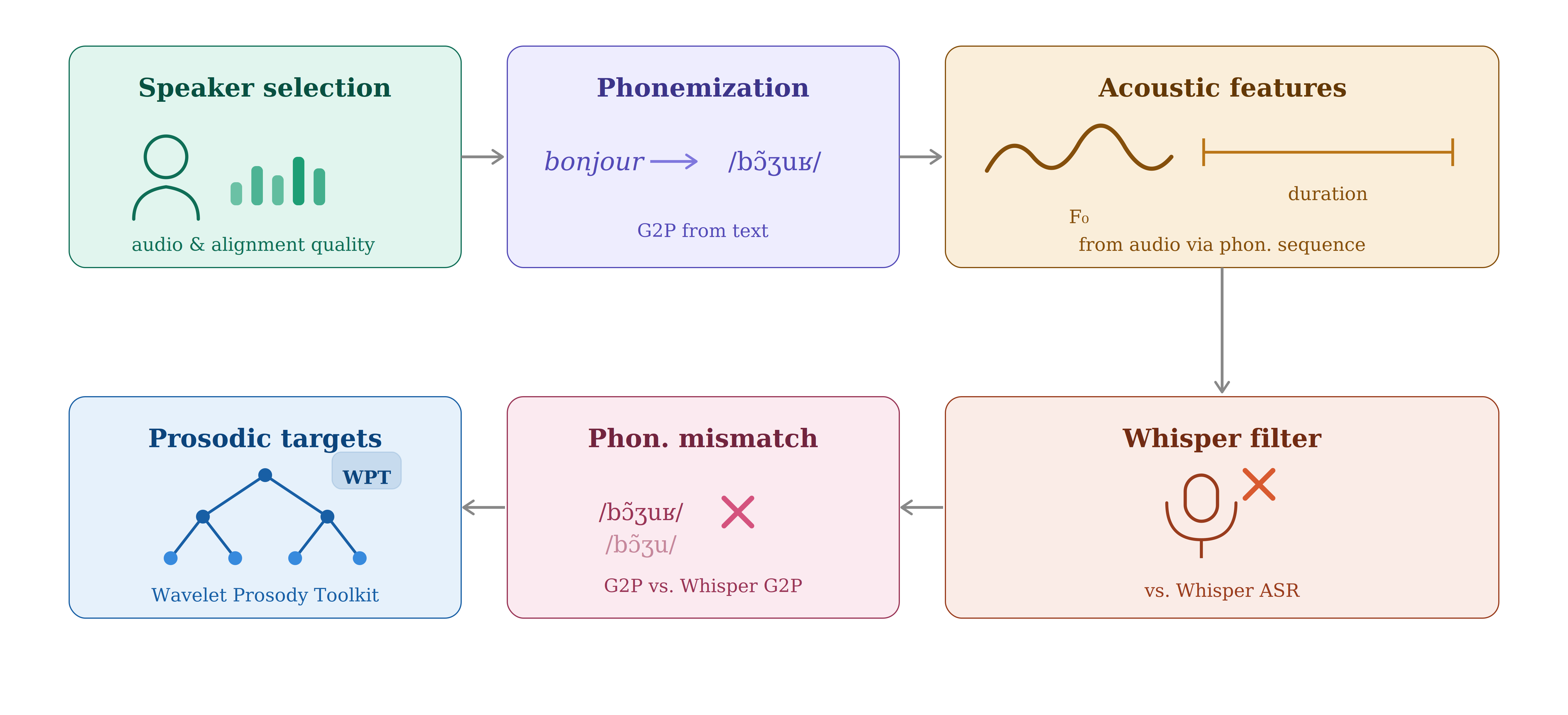}
\caption{Pre-processing workflow applied to our dataset}
\label{data_pipeline}
\end{figure}

We experiment on a subset of a multi-speaker corpus of open-source French audiobooks called \textbf{Mufasa} \cite{Mufasa}. 
As shown in Figure~\ref{data_pipeline}, the curation of this subset for TTS experimentations includes various steps. We start by selecting speakers based on the global quality of their recording. %
Since the recorded volume is uneven across the 9 chosen speakers (5 women and 4 men), almost a third of our subset is attributed to two female speakers.

In the following steps, the phonemized syllable sequences needed for the guidance predictor (see Section~\ref{bipp_guidance}) are generated alongside segment duration and fundamental frequency values, using an in-house toolkit. 
Sample pairs whose text does not correspond to Whisper's transcription \cite{radford2022robustspeechrecognitionlargescale} are filtered out. 
The \textbf{continuous word-level prosodic targets} are then computed on the remaining audio samples with the help of a \textbf{modified version of the WPT} (in which the audio features are normalized over the dataset rather than the individual audio sample), and broadcasted on syllables during the final step. These modifications should not affect the correspondence between WPT values and human prosodic perception, established by Suni et al.~\cite{SUNI_2017_wavelet}. We also discard the original classification step of the toolkit in favor of our own.
After randomly isolating 10k samples to split between the development and the test subsets equally, we are left with approximately 47k training samples ($\sim$ 45 hours) from the curated 57k pairs.

\section{Experimental Setting}\label{expe_setup}

Exception made of \vI~(which required an additional 100k steps), all TTS models are trained up to 500k steps on the previously described \textit{train} dataset, with a batch size of 64. Training is stopped based on the evaluation results on the \textit{development} data. 
The BIPP front-end model only requires 50k steps based on the results on the \textit{development} set. 
We make it estimate prominence and boundary values for all transcriptions of the \textit{test} set, to serve as external guidance in our experiments.

The same texts are synthesized under 3 inference configurations: a) guidance-free; b) BIPP-guided; c) oracle-guided. In the oracle-guided configuration, the systems are given information extracted from the reference recordings (i.e. ground truth prominences and boundaries).
In the case of TTS models deprived of dedicated predictors only the last 2 configurations are used to synthesize audio. 

We compute prominences and boundaries for all synthetic signals using our modified version of the WPT \cite{SUNI_2017_wavelet}, to facilitate comparison with ground truth samples and guidance sequences. 
The labels sequences are obtained by applying thresholds on each continuous prosodic event sequence, as in \cite{mas-etal-2026-automatic}. 
Continuous prominence values are thus converted into "None", "Medium" or "Strong" labels, where "Strong" is attributed to the 10\% highest values of the dataset distribution, and "Medium" expands this range to 20\%.
The same method is applied to obtain boundary labels. 

\section{Evaluation}\label{eval}

Our investigations of prosodic modulation in speech synthesis can be divided into three stages. 
We first assess system-wise performance in terms of prosody faithfulness, across the 3 configurations (Section~\ref{expe1}). 
We then turn our focus toward the controllability of our systems in Section~\ref{expe2}, where we further investigate whether BIPP's instructions are properly followed, and whether one input format is better suited for this task. 
Finally, we assess the impact of prosodic events on synthesis quality using UTMOS (Section~\ref{expe3}), an independent metric that does not rely on the modified WPT.

As noted in the introduction \cite{mas-etal-2026-automatic}, salient prosodic events are particularly relevant for the rendering of a speaker's prosody.
Hence, we introduce per-label classification metrics to the first two stages of investigation, in addition to regression-related metrics.

\subsection{Results in Prominence and Boundary Restitution}\label{expe1}

To get an idea of each system's performance and the impact of the different guidance types, we conduct a first study where we compare synthetic speech with the natural recordings at disposal, using the \textit{test} set previously mentioned. We successively compute a set of metrics on the event sequences extracted for every reference sample and each of its synthetic counterparts. More specifically, we compute, for prominences and boundaries, i) the \textbf{Mean Squared Error (MSE)} and \textbf{Dynamic Time Warping (DTW)} reported in Table~\ref{tab:compare_gt_cont}, as well as ii) a \textbf{per-class averaged F1 score} reported in Table~\ref{tab:compare_gt_classif_sym03}.

\definecolor{rowblue}{rgb}{0.85,0.92,1.0}
\definecolor{bestcol}{HTML}{1a7a1a}
\definecolor{badcol}{HTML}{c0392b}

\newcommand{\imp}[1]{\textcolor{bestcol}{\textbf{#1}}}
\newcommand{\worse}[1]{\textcolor{badcol}{\textit{#1}}}

\begin{table}[!t]
\centering
\caption{Comparison between synthesized and natural speech in terms of prominence and boundaries values}
\resizebox{\linewidth}{!}{%
    \begin{tabular}{ll cccc}
    \toprule
    \multirow{2}{*}{\textbf{System}}
    & \multirow{2}{*}{\textbf{Guidance}}
      & \multicolumn{2}{c}{\textbf{Prominences}}
      & \multicolumn{2}{c}{\textbf{Boundaries}} \\
    \cmidrule(lr){3-4}\cmidrule(lr){5-6}
     & & \textit{MSE}$\downarrow$ & \textit{DTW}$\downarrow$
       & \textit{MSE}$\downarrow$ & \textit{DTW}$\downarrow$ \\
    \midrule
    \texttt{FS base}    & ---
      & 0.22 & 4.89 & 0.17 & 3.31 \\
    \midrule
    \multicolumn{6}{l}{\textit{No BIPP guidance}} \\
    \midrule
    \rowcolor{rowblue}
    \texttt{FS prom} & ---
      & \worse{0.26}$^b$ & \imp{3.81}$^b$ & \imp{0.15}$^b$ & \imp{2.39}$^b$ \\
    \texttt{\vI}              & ---
      & \worse{0.24}$^{b,w}$ & \imp{3.62}$^{b,w}$ & \imp{0.11}$^{b,w}$ & \imp{2.14}$^{b,w}$ \\
    \texttt{\vII}        & ---
      & \worse{0.24}$^{b,w}$ & \imp{3.65}$^{b,w}$ & \imp{0.11}$^{b,w}$ & \imp{2.14}$^{b,w}$ \\
    \rowcolor{rowblue}
    \texttt{\vIII} & ---
      & \imp{0.14}$^{b,w}$ & \imp{3.02}$^{b,w}$ & \imp{0.11}$^{b,w}$ & \imp{2.11}$^{b,w}$ \\
    \midrule
    \multicolumn{6}{l}{\textit{BIPP guidance}} \\
    \midrule
    \texttt{FS prom}          & BIPP
      & \worse{0.26}$^b$ & \imp{3.87}$^b$ & \imp{0.16}$^{b,w}$ & \imp{2.47}$^{b,w}$ \\
    \texttt{\vI}              & BIPP
      & \worse{0.25}$^b$ & \imp{3.84}$^b$ & \imp{0.15}$^b$ & \imp{2.42}$^b$ \\
    \texttt{\vII}        & BIPP
      & \worse{0.26}$^b$ & \imp{3.86}$^b$ & \imp{0.15}$^b$ & \imp{2.42}$^b$ \\
    \texttt{\vIII} & BIPP
      & \worse{0.26}$^b$ & \imp{3.90}$^b$ & \imp{0.15}$^b$ & \imp{2.42}$^b$ \\
    \rowcolor{rowblue}
    \texttt{FS labels} & BIPP
      & \worse{0.26}$^b$ & \imp{3.81}$^b$ & \imp{0.13}$^{b,w}$ & \imp{2.29}$^{b,w}$ \\
    \midrule
    \multicolumn{6}{l}{\textit{Oracle guidance}} \\
    \midrule
    \texttt{FS prom}          & Oracle
      & \imp{0.20}$^{b,w}$ & \imp{3.60}$^{b,w}$ & \imp{0.14}$^{b,w}$ & \imp{2.39}$^b$ \\
    \texttt{\vI}              & Oracle
      & \imp{0.20}$^{b,w}$ & \imp{3.59}$^{b,w}$ & \imp{0.12}$^{b,w}$ & \imp{2.25}$^{b,w}$ \\
    \texttt{\vII}        & Oracle
      & \imp{0.20}$^{b,w}$ & \imp{3.57}$^{b,w}$ & \imp{0.12}$^{b,w}$ & \imp{2.24}$^{b,w}$ \\
    \texttt{\vIII} & Oracle
      & \imp{0.17}$^{b,w}$ & \imp{3.29}$^{b,w}$ & \imp{0.12}$^{b,w}$ & \imp{2.20}$^{b,w}$ \\
    \rowcolor{rowblue}
    \texttt{FS labels} & Oracle
      & \imp{0.15}$^{b,w}$ & \imp{3.02}$^{b,w}$ & \imp{0.11}$^{b,w}$ & \imp{2.10}$^{b,w}$ \\
    \bottomrule
    \multicolumn{6}{l}{\textit{Significance computed against FS base's and FS prom's metrics \footnotemark: }} \\
    \multicolumn{6}{l}{\textit{- $^b$ : 95\% bootstrap CI non-overlapping with \texttt{FS base};}} \\
    \multicolumn{6}{l}{\textit{- $^w$ : also non-overlapping with \texttt{FS prom}; }} \\
    \multicolumn{6}{l}{\textit{\imp{Green bold} / \worse{red italic} : \textit{\imp{improvements}}/\worse{regressions} relatively to \texttt{FS base}.}} \\
    \end{tabular}%
}
\label{tab:compare_gt_cont}
\end{table}
\footnotetext{\label{fn:shaded}Blue-shaded rows indicate models of particular interest.}

Taking a first look at Table~\ref{tab:compare_gt_cont}, we notice that all systems are significantly different from FS base (our baseline) with respect to ground truth continuous events. 
Interestingly, boundary values are systematically closer to the expectations as long as the systems have a notion of prosodic events, which is not the case for prominence values. 
Indeed, the only improvements reported for the prominence's MSE are attributed to the \vIII~variant and the oracle-based configurations. 
The \vIII~variant happens to see oracle-derived labels for both prominence and boundaries in addition to the intermediate continuous predictors, and it is all the more interesting to see that it outperforms even FS labels (deprived of intermediate prosodic predictions) in almost every metric. %

Turning our attention to the whole oracle-guidance sub-table, we observe quite logically that this configuration significantly improves all metrics compared to both FS base and FS prom (without guidance), exception made of the boundary DTW for the guided FS prom. FS labels and \vIII~perform best in this section. Although they all still fall behind \vIII~in its default configuration, they are visibly closer to the ground truth compared to the "BIPP guidance" configurations. 

Speaking of BIPP, its guidance seems to have influence on the results as well. For a start, all metrics are significantly different to FS base when BIPP's sequences are fed into the other systems. The regressions on prominences and improvements on boundaries observed here seem coherent with the conclusions drawn by Mas et al. \cite{mas-etal-2026-automatic} on the performance of BIPP itself, compared to their FastSpeech baseline. Moreover, only the boundaries metrics of a couple of systems are significantly improved compared to FS prom: FS prom (with BIPP guidance) and FS labels. It seems that combining input labels and intermediate predictors do not lead to further improvement compared to either of the two in this configuration. 

\definecolor{rowblue}{rgb}{0.85,0.92,1.0}
\definecolor{bestcol}{HTML}{1a7a1a}
\definecolor{badcol}{HTML}{c0392b}

\begin{table}[!ht]
\centering
\caption{Comparison between synthesized and natural speech in terms of  prominence and boundary labels (per class F1 score)}
\resizebox{\linewidth}{!}{%
    \begin{tabular}{ll ccc ccc}
        \toprule
        \multirow{2}{*}{\textbf{System}}
        & \multirow{2}{*}{\textbf{Guidance}}
          & \multicolumn{3}{c}{\textbf{Prominences}}
          & \multicolumn{3}{c}{\textbf{Boundaries}} \\
        \cmidrule(lr){3-5}\cmidrule(lr){6-8}
         & & \textit{None} & \textit{Med.} & \textit{Strong}
           & \textit{None} & \textit{Med.} & \textit{Strong} \\
        \midrule
        \texttt{FS base}    & ---
          & 0.89 & 0.29 & 0.35
          & 0.87 & 0.21 & 0.56 \\
        \midrule
        \multicolumn{8}{l}{\textit{No BIPP guidance}} \\
        \midrule
        \texttt{FS prom} & ---
          & 0.87 & 0.29 & \imp{0.40}
          & 0.87 & 0.20 & \imp{0.66} \\
        \texttt{\vI}              & ---
          & 0.88 & 0.30 & \imp{0.44}
          & \imp{0.91} & \imp{0.31} & \imp{0.73} \\
        \texttt{\vII}        & ---
          & 0.88 & 0.30 & \imp{0.45}
          & \imp{0.91} & \imp{0.31} & \imp{0.73} \\
        \rowcolor{rowblue}
        \texttt{\vIII} & ---
          & \imp{0.94} & \imp{0.46} & \imp{0.67}
          & \imp{0.91} & \imp{0.31} & \imp{0.73} \\
        \midrule
        \multicolumn{8}{l}{\textit{BIPP guidance}} \\
        \midrule
        \texttt{FS prom}          & BIPP
          & 0.87 & 0.30 & \imp{0.41}
          & 0.87 & \worse{0.18} & \imp{0.67} \\
        \texttt{\vI}              & BIPP
          & 0.87 & 0.30 & \imp{0.40}
          & 0.88 & \worse{0.18} & \imp{0.69} \\
        \texttt{\vII}        & BIPP
          & 0.87 & 0.29 & \imp{0.39}
          & 0.88 & \worse{0.18} & \imp{0.68} \\
        \texttt{\vIII} & BIPP
          & 0.87 & 0.26 & \imp{0.38}
          & 0.88 & \worse{0.16} & \imp{0.69} \\
        \rowcolor{rowblue}
        \texttt{FS labels} & BIPP
          & 0.88 & 0.27 & 0.37
          & 0.89 & 0.19 & \imp{0.70} \\
        \midrule
        \multicolumn{8}{l}{\textit{Oracle guidance}} \\
        \midrule
        \texttt{FS prom}          & Oracle
          & 0.89 & \imp{0.34} & \imp{0.48}
          & 0.88 & 0.22 & \imp{0.68} \\
        \texttt{\vI}              & Oracle
          & 0.89 & \imp{0.35} & \imp{0.50}
          & \imp{0.91} & \imp{0.28} & \imp{0.72} \\
        \texttt{\vII}        & Oracle
          & 0.89 & \imp{0.34} & \imp{0.50}
          & \imp{0.91} & \imp{0.28} & \imp{0.72} \\
        \texttt{\vIII} & Oracle
          & \imp{0.93} & \imp{0.44} & \imp{0.64}
          & \imp{0.91} & \imp{0.29} & \imp{0.72} \\
        \rowcolor{rowblue}
        \texttt{FS labels} & Oracle
          & \imp{0.94} & \imp{0.45} & \imp{0.66}
          & \imp{0.91} & \imp{0.30} & \imp{0.73} \\
        \bottomrule
        \multicolumn{8}{l}{} \\
        \multicolumn{8}{l}{\textit{Colors indicate noteworthy differences by magnitude relative to \texttt{FS base}:}} \\
        \multicolumn{8}{l}{\textit{ \imp{Green bold} $\geq+0.03$, \worse{red italic} $\leq-0.03$. \footnotemark[\getrefnumber{fn:shaded}]}} \\
    \end{tabular}%
}
\\[3pt]
\label{tab:compare_gt_classif_sym03}
\end{table}

\definecolor{rowblue}{rgb}{0.85,0.92,1.0}
\definecolor{bestcol}{HTML}{1a7a1a}
\definecolor{badcol}{HTML}{c0392b}
\definecolor{bestcol2}{HTML}{1a52a8}
\definecolor{badcol2}{HTML}{8e44ad}

\newcommand{\impC}[1]{\textcolor{bestcol2}{\textbf{#1}}}
\newcommand{\worseC}[1]{\textcolor{badcol2}{\textit{#1}}}

\begin{table*}[ht]
\centering
\caption{Regression- and Classification-based evaluation of synthetic speech against BIPP's guidance}
\resizebox{0.7\linewidth}{!}{%
    \begin{tabular}{l cccc cc cc}
    \toprule
    \multirow{3}{*}{\textbf{System (all BIPP)\textsuperscript{*}}}
      & \multicolumn{4}{c}{\textbf{Continuous}}
      & \multicolumn{4}{c}{\textbf{Per-class F1}} \\
    \cmidrule(lr){2-5}\cmidrule(lr){6-9}
      & \multicolumn{2}{c}{\textit{Prominences}}
      & \multicolumn{2}{c}{\textit{Boundaries}}
      & \multicolumn{2}{c}{\textit{Prominence}}
      & \multicolumn{2}{c}{\textit{Boundaries \textsuperscript{†}}} \\
    \cmidrule(lr){2-3}\cmidrule(lr){4-5}\cmidrule(lr){6-7}\cmidrule(lr){8-9}
      & \textit{Corr}$\uparrow$ & \textit{MSE}$\downarrow$
      & \textit{Cos.\ Sim.}$\uparrow$ & \textit{MSE}$\downarrow$
      & \textit{Med F1}$\uparrow$ & \textit{Strong F1}$\uparrow$
      & \textit{None F1}$\uparrow$ & \textit{Med F1}$\uparrow$ \\
    \midrule
    \multicolumn{9}{l}{\textit{Baselines}} \\
    \midrule
    \rowcolor{gray!10}
    \texttt{FS base}
      & 0.72 & 0.61 & 0.353 & 1.586
      & 0.17 & 0.29 & 0.88 & 0.44 \\
    \rowcolor{gray!10}
    \texttt{FS prom} (no guidance)
      & 0.77 & 0.66 & 0.151 & 1.203
      & 0.19 & 0.32 & 0.91 & 0.51 \\
    \midrule
    \multicolumn{9}{l}{\textit{Labels-agnostic model}} \\
    \midrule
    \texttt{FS prom + BIPP}
      & \imp{0.82}$^{b,w}$ & \imp{0.58}$^{b,w}$ & 0.155$^{b}$ & \imp{1.151}$^{b,w}$
      & 0.21 & 0.40 & 0.92 & 0.52 \\
    \midrule
    \multicolumn{9}{l}{\textit{Label models}} \\
    \midrule
    \texttt{\vI}
      & 0.83$^{b,w}$ & \impC{0.30}$^{b,w,c}$ & \impC{0.994}$^{b,w,c}$ & \impC{0.005}$^{b,w,c}$
      & 0.22 & 0.41 & 0.93 & 0.56 \\
    \texttt{\vII}
      & 0.82$^{b,w}$ & \impC{0.31}$^{b,w,c}$ & \impC{0.994}$^{b,w,c}$ & \impC{0.005}$^{b,w,c}$
      & 0.21 & 0.40 & 0.93 & 0.55 \\
    \rowcolor{rowblue}
    \texttt{\vIII}
      & \worseC{0.81}$^{b,w,c}$ & \impC{0.29}$^{b,w,c}$ & \impC{0.994}$^{b,w,c}$ & \impC{0.005}$^{b,w,c}$
      & 0.27 & 0.54 & 0.94 & 0.59 \\
    \rowcolor{rowblue}
    \texttt{FS labels}
      & \worseC{0.79}$^{b,w,c}$ & \impC{0.31}$^{b,w,c}$ & \impC{0.995}$^{b,w,c}$ & \impC{0.004}$^{b,w,c}$
      & 0.28 & 0.57 & 0.94 & 0.59 \\
    \bottomrule
    \multicolumn{9}{l}{} \\
    \multicolumn{9}{l}{\textit{$^b$: significant (sig.) vs.\ \texttt{FS base};\quad $^w$: sig. vs.\ \texttt{FS prom} (no guidance); \quad $^c$: sig. vs.\ \texttt{FS prom + BIPP}.}} \\
    \multicolumn{9}{l}{\textit{\imp{green bold}: sig. \textit{\imp{improvement}} vs.\ \texttt{FS prom} (no guidance) for the \texttt{FS prom BIPP} row only.}} \\
    \multicolumn{9}{l}{\textit{\impC{blue bold}/\worseC{purple italic}: sig \textit{\impC{improvements}}/\worseC{regressions} vs.\ \texttt{FS prom + BIPP} (label model rows, continuous metrics).\footnotemark[\getrefnumber{fn:shaded}]}} \\
    \multicolumn{9}{l}{\textsuperscript{*} \textit{All values are re-extracted by the modified WPT from the \emph{synthesized} audio.}} \\ %
    \multicolumn{9}{l}{\textsuperscript{†} \textit{Bound.\ Strong~F1 = 0: BIPP never requests Strong boundaries, so control is shown over the None/Medium range only.}} \\ %
    
    \end{tabular}%
}
\label{tab:compare_guidance}
\end{table*}

Looking now at the classification metrics in Table~\ref{tab:compare_gt_classif_sym03}, we observe that only the identification of medium boundaries in the BIPP-guidance setting decreases by more than a 0.03 margin (with the exception of FS labels). 
The difference in coloring compared to the previous table explains itself a) by the chosen coloring thresholds and b) by the fact that 80\% of the values are gathered under the "None" labels for both event types. 
The strong prominence and boundary labels globally benefit from prosodic awareness in the models, while the medium labels show less variation.
Again \vIII~(not guided) and FS labels display good performance compared to the other models.  

The greater gaps in performance on salient event classes (medium and strong) depending on the guidance quality suggest that the label-based model allow for more controllability compared to FS prom. \vIII~and FS labels responses are particularly strong in comparison to the other systems. To confirm this claim, we change the evaluation reference from ground truth speech to BIPP's guidance in the next section.

\subsection{Prosodic Control}\label{expe2}

Prominences and boundaries, as human-interpretable representations derived from physical measurements, are presumably well-suited for prosody control. 
In order to quantify the extent to which this control is achieved across our systems, we compare this time the original guidance with the synthesized speech of each system receiving BIPP's guidance. 
We choose to focus on this configuration since BIPP's guidance, predicted solely from enriched textual representation, exhibits different characteristics to the ground-truth based oracles, or internal predictions of the TTS models. 
As a matter of fact, it did not predict boundary values strong enough to be categorized as "strong" for any \textit{test} text samples, thus the absence of this label in our results. 
FS base and FS prom are also included in this experiment to control the extent to which their generated speech already matches the guidance given in other synthesis configurations.

We compute \textbf{Pearson correlation}, \textbf{cosine similarity}, and \textbf{MSE} between continuous prosodic events given as guidance and those retrieved from synthetic speech. 
For prominences, all three metrics agree closely, so Table~\ref{tab:compare_guidance} reports only Pearson correlation and MSE. 
Since boundary sequences are sparser (mostly near-zero, with a few spikes), we report the better-suited cosine similarity.
Categorical metrics are also computed, and we report mean per-class F1 scores for conciseness.

As shown in the results compiled in Table~\ref{tab:compare_guidance}, FS Base and FS prom already show a somehow strong correlation to the guidance in terms of prominence, almost reaching 0.8 in the case of FS prom. 
However, they are quite distant in terms of boundaries, although they do exhibit higher F1 scores for this type of events.  
When given the actual guidance, FS prom significantly closes the gap on prominence regression metrics, and manages to improve its F1 scores as well. 
It still struggles with boundaries, for which not all comparison yield significant difference.  

The label-based models, on the other hand, all benefit from the boundary labels they receive and greatly improve the similarity to the guidance for this type of events. 
Interestingly, we obtain mixed performance for prominences in comparison with FS prom, when looking at systems which do receive them as labels (namely \vIII~and FS labels). 
While the MSE drops from 0.58 to 0.3 approximately, the correlation actually suffers a regression.
This is confirmed by intra-label system comparisons, not marked in the table. 
Classification metrics also show some improvement, especially for strong prominences.

Overall, label-based systems are very competitive against FS prom when comes guidance-following. 
FS labels sticks out as the weakest in following prominence instructions when come regression metrics, which contrasts with the story told by classification metrics. 
When evaluating salient prosodic events, the performance of this system appears to be slightly less impacted by the defined threshold for medium and strong events than the others.
It is worth noting that near-perfect directional alignment globally does not translate to near-perfect class attribution. A possible explanation for this is the impact of the thresholding step used to obtain the labels.

\definecolor{sigbetter}{RGB}{198,239,206}   %
\definecolor{sigworse} {RGB}{255,199,206}   %
\definecolor{refrow}   {RGB}{230,230,245}   %
\definecolor{gtrow}    {RGB}{245,245,245}   %
\definecolor{fyi}      {RGB}{220,235,255}   %

\newcommand{\up}  [1]{\textcolor{green!50!black}{$\uparrow$\,#1}}
\newcommand{\down}[1]{\textcolor{red!70!black}{$\downarrow$\,#1}}
\newcommand{\ns}      {\textcolor{gray}{--}}

\begin{table}[ht]
\centering
\caption{Estimated marginal means of UTMOSv2's scores }
\setlength{\tabcolsep}{5pt}
\begin{adjustbox}{max width=\linewidth}
    \begin{tabular}{llc ccc}
      \toprule
      \textbf{Group} & \textbf{System} & \textbf{EMM}
        & \emph{vs.}~\textsc{fs base}\textsuperscript{*}
        & \emph{vs.}~\textsc{fs prom}\textsuperscript{*}
        & \emph{vs.}~\textsc{fs prom bipp}\textsuperscript{*} \\
      \midrule
      \rowcolor{gtrow}
      Ref.\ (natural) & Ground Truth                  & 3.02 & \up{$+$0.11} & \up{$+$0.11} & \up{$+$0.17} \\
      \midrule
      \rowcolor{refrow}
      Baseline & \textsc{fs base}                      & 2.91 & \ns           & \ns           & \up{$+$0.06} \\
      \rowcolor{refrow}
               & \textsc{fs prom}                      & 2.91 & \ns           & \ns           & \up{$+$0.05} \\
      \midrule
      \multirow{5}{*}{BIPP}
        & \textsc{fs prom bipp}                        & 2.85 & \down{$-$0.06} & \down{$-$0.05} & \ns             \\
        & \textsc{fsl1 bipp}                           & 2.88 & \down{$-$0.03} & \down{$-$0.03} & \up{$+$0.03}   \\
        & \textsc{fsl2 bipp}                           & 2.88 & \down{$-$0.03} & \down{$-$0.03} & \up{$+$0.02}   \\
        & \textsc{fsl3 bipp}                           & 2.87 & \down{$-$0.04} & \down{$-$0.04} & \ns             \\
      \rowcolor{sigbetter}
        & \textsc{fs labels bipp}                      & \textbf{2.94} & \up{$+$0.03} & \up{$+$0.03} & \up{$+$0.08} \\
      \midrule
      \multirow{4}{*}{Oracle}
        & \textsc{fsl1 oracle}                         & 2.88 & \down{$-$0.03}           & \down{$-$0.03} & \up{$+$0.03} \\
        & \textsc{fsl2 oracle}                         & 2.89 & \down{$-$0.02}$^\dagger$ & \ns             & \up{$+$0.04} \\
        & \textsc{fsl3 oracle}                         & 2.89 & \down{$-$0.02}$^\dagger$ & \ns             & \up{$+$0.04} \\
       \rowcolor{refrow} %
        & \textsc{fs labels}                           & 2.91 & \ns           & \ns           & \up{$+$0.06} \\
      \midrule
      \multicolumn{2}{l}{\textit{\textsc{fsl1}--\textsc{fsl3}}} & \textit{2.90--2.91} & \ns & \ns & \up{$+$0.05--$+$0.06} \\
      \multicolumn{6}{l}{\textit{(no guidance): all n.s.\ vs.\ baselines}} \\
      \bottomrule
      \multicolumn{6}{l}{} \\
      \multicolumn{6}{l}{\textsuperscript{*}\textit{Cells show $\Delta$ EMM with direction: \color{green!50!black}$\uparrow$\,better ; \color{red!70!black}$\downarrow$\,worse}. $\dagger$: marginally significant ($p<0.05$).}  \\
    \end{tabular}
\end{adjustbox}
\label{tab:utmos_condensed}
\end{table}

\subsection{Overall Synthesis Quality}\label{expe3} 

\begin{figure*}[htpb]
\centering
\includegraphics[width=0.8\textwidth]{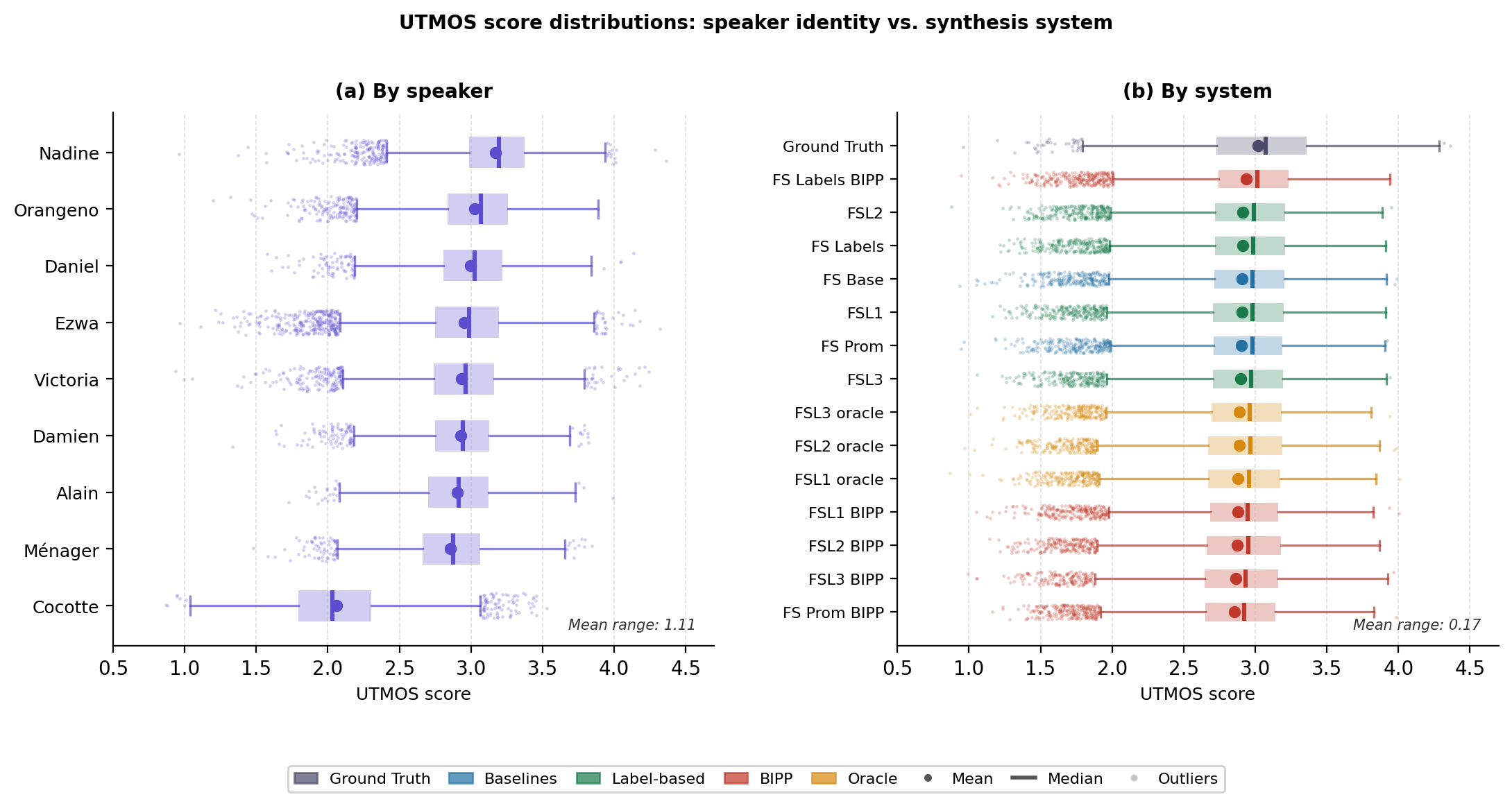}
\caption{Boxplot of UTMOSv2' estimated MOS scores grouped (a) by speaker (collapsed across systems) and (b) by system (collapsed across speakers)}
\label{utmos_groupings}
\end{figure*}

As a first estimate of human judgments of naturalness, we compute \textbf{Mean Opinion Scores (MOS)} predictions per audio with UTMOSv2 \cite{baba_utmos_2024} and average it over speakers and systems. 
Also computed for reference, the mean estimated MOS of natural speech from our dataset is 3.02.  

The average scores of all configurations fall close to this reference, varying between 2.85 and 2.95 approximately. 
Considering how limited this range is, we first perform a two-way ANOVA on the estimated score, suspecting that the speaker variable may explain a significant part of the variability in our results. 
As it turns out, the speaker variable accounts for 90.3\% of this variability, whereas our systems only account for 1.3\% (residual of 8.4). 
This is well illustrated by the two boxplots in figure \ref{utmos_groupings}.
To properly compare the estimated scores across systems in this context, we turn to a Linear Mixed Model (LMM) \cite{bates2015lme4, lenth2016lsmeans} and treat the per-sample effect as "random". %

The \textbf{Estimated Marginal Means (EMM)} produced by the LMM after averaging out the random effects are reported in Table~\ref{tab:utmos_condensed}. 
The model variants \vI, \vII~and \vIII~do not exhibit an EMM significantly different from either FS base or FS prom. 
Adding prosodic labels as input therefore does not hurt the estimated naturalness in a standard setup, though we can also say that it does not improve it.  
However, most systems significantly outperform FS prom when all are fed with BIPP guidance, including \vI~and \vII~variants.  %
They also close the gap with FS prom compared the same model fed with BIPP guidance, suggesting that label do help when external guidances are involved.
Interestingly, oracle-guided systems do not improve the scores compared to the two reference systems. 
We don't observe significant differences at best, and even a small (but significant) degradation at worse. 
Brought in contrast with the results from Section~\ref{expe1}, this highlights that a) label instructions do introduce variability \textbf{without impacting the estimated naturalness} and b) prosodic appropriateness is not a strong criterion in UTMOS's assessments.

\section{Conclusion}

In this work, we compared different strategies to add prosodic control in a FastSpeech2 model using prominence and boundary events as a representation. 
We developed several TTS models trained with either prosodic labels, dedicated prosody predictors, or both.  
Our experiments show that prosody-awareness globally improve FastSpeech's performance relative to expected prosody. Having knowledge of prominences and/or boundaries indeed reduces the prosodic distance with natural speech for most of the metrics. 
This is consistent with the findings of Suni et al. \cite{suni_prosodic_2020} on a Tacotron model.
When comes modulation, external guidance significantly impacts the generated prosody, regardless of if it comes from a front-end predictor or from an oracle. 
These results point towards a better controllability of the label-informed models compared to the model relying solely on co-trained continuous predictors.
Our mechanism allows TTS systems to be realigned with user expectations without changing the synthesis model, which we hypothesize could benefit perceived quality.
As mentioned above, UTMOS aims to predict naturalness from purely acoustic criteria and shows its limitations in evaluating the perceived quality since it fails at discriminating between prosodic conditioning strategies.
A human listening test would be needed to attest the impact of our method in terms of prosody appropriateness.  

\section{Acknowledgments}

The formatting of result tables was assisted by Claude Sonnet 4.6 and Opus 4.8. The same systems served as linguistic assistants for the first half of our introduction.
This work was performed using HPC resources from GENCI-IDRIS (Grant 2025-AD011016834).
We would like to thank Laure Charonnat (Voxygen) for her technical support and helpful advice. %

\bibliographystyle{IEEEtran}
\bibliography{ref}

\end{document}